\documentclass{article}

\usepackage{arxiv}

\usepackage[utf8]{inputenc} % allow utf-8 input
\usepackage[T1]{fontenc}    % use 8-bit T1 fonts
\usepackage{hyperref}       % hyperlinks
\usepackage{url}            % simple URL typesetting
\usepackage{booktabs}       % professional-quality tables
\usepackage{amsfonts}       % blackboard math symbols
\usepackage{nicefrac}       % compact symbols for 1/2, etc.
\usepackage{microtype}      % microtypography
\usepackage{lipsum}		% Can be removed after putting your text content
\usepackage{amssymb}
\usepackage{amsmath}
\usepackage{graphicx}
\usepackage{doi}
\usepackage{xspace}

\usepackage{multirow,siunitx}
\usepackage{diagbox}
\usepackage[table]{xcolor}

\usepackage[
backend=biber,
style=authoryear,
sorting=ynt
]{biblatex}
\title{Sectioning of Biomedical Abstracts: A Sequence of Sequence Classification Task}

\author{{Mehmet Efruz Karabulut} \\
    Computer \& Information Sciences \\
    University of Delaware \\ 
    Newark, DE, USA \\ 
	\And
	{K. Vijay-Shanker} \\
	Computer \& Information Sciences \\
    University of Delaware \\ 
    Newark, DE, USA \\ 
}

\hypersetup{
pdftitle={A template for the arxiv style},
pdfsubject={q-bio.NC, q-bio.QM},
pdfauthor={David S.~Hippocampus, Elias D.~Striatum},
pdfkeywords={First keyword, Second keyword, More},
}

\begin{document}
\maketitle

\begin{abstract}
	Rapid growth of the biomedical literature has led to many advances in the biomedical text mining field. Among the vast amount of information, biomedical article abstracts are the easily accessible sources. However, the number of the structured abstracts, describing the rhetorical sections with one of Background, Objective, Method, Result and Conclusion categories is still not considerable. Exploration of valuable information in the biomedical abstracts can be expedited with the improvements in the sequential sentence classification task. Deep learning based models has great performance/potential in achieving significant results in this task. However, they can often be overly complex and overfit to specific data. In this project, we study a state-of-the-art deep learning model, which we called SSN-4 model here. We investigate different components of the SSN-4 model to study the trade-off between the performance and complexity. We explore how well this model generalizes to a new data set beyond Randomized Controlled Trials (RCT) dataset. We address the question that whether word embeddings can be adjusted to the task to improve the performance. Furthermore, we develop a second model that addresses the confusion pairs in the first model. Results show that SSN-4 model does not appear to generalize well beyond RCT dataset.
\end{abstract}

% keywords can be removed

\section{Introduction}

Over the past years, there has been a rapid increase in the biomedical literature. As of today, over 30 million biomedical papers have been indexed through PubMed. Accordingly, the growth in the biomedical literature has resulted in considerable interest in the biomedical text mining. While this trend promises new research directions, it simultaneously brings challenges towards taking advantage of all the relevant information due to its vast amount. 

Most biomedical text mining research is concerned with the biomedical article abstracts. They are freely available via Medline and they provide the key information in the articles. Nevertheless, most of the abstracts are unstructured which means they don't have semantic headings such as Background, Method, Result, and Conclusion. Extraction of desired information from unstructured abstracts often doesn't include information about the section it appears in. Structured abstracts alleviate the need to access divergent biomedical literature to locate relevant articles and read them to locate relevant information. For instance, identification of semantic headings in abstracts can help us gain valuable knowledge in exploration, e.g., which information is a result of an experimental finding reported in a paper, what text might provide text evidence for extracted relations, or for summarization of genes/proteins. 

In a scientific abstract each sentence can be assigned to a rhetorical element. This task has also called sequential sentence classification. The assignment of each sentence is greatly associated with the assignment of other sentences in the abstract. For example assigning the role "result" to a sentence indicates that the next element is most likely to be either a "result" or a "conclusion" sentence. Preceding studies on the task mainly focused on feature-based Machine Learning methods such as naive Bayes [Huang et al., 2013], Support Vector Machines [Liu et al., 2013] or traditional sequence labeling models such as Hidden Markov Model [Lin et al., 2006] and Conditional Random
Fields [Kim et al., 2011]. These models require feature engineering based on lexical, semantic or
sequential information to classify sentences. In recent years Deep Neural Networks (DNNs) based models are designed to automatically learn the features based on the embeddings. However, the problem with many of those models is that they classify sentences in isolation, without considering the contextual information. Recently, [Jin
and Szolovits, 2018] proposed a model that combines the representational power of deep learning
with the hierarchical and sequential nature of the abstracts and achieved state-of-the-art results. 

In this project, we study the model of [Jin and Szolovits, 2018] in detail. The model is fairly complex and it has several layers trying to capture different aspects of the problem. Here we will investigate the contribution of the components of a complex model. We address the question that whether lexical representation can be tailored to the task. We explore the ability of the model to generalize beyond the authors' annotated set. Based on analysis we present an approach to improve the accuracy of the model.

\section{Related Work}
Before the emergence of Deep Neural Networks, a large body of work on sentence classification task manually crafted features by experimenting on the structural and the semantic relations in the task-specific dataset. [\cite{ruch2007using}] used Naive Bayes classifiers to distinguish the key sentence in a biomedical abstract. They used N-grams and position heuristics of those segments as features. In a similar task, 
[\cite{liu2013abstract}] based their sentence classifier on Support Vector Machine (SVM) that classifies abstract sentences into four categories. Furthermore, they proposed a semi-supervised learning method which is Transductive Support Vector Machine (TSVM) to mitigate the data scarcity problem. In both studies sentence classification happens in isolation, therefore the sequential order of the predicted categories is not considered. Example of works that benefited from sequential features include traditional sequence labeling models such as Hidden Markov Model [\cite{lin2006generative}] and Conditional Random Fields [\cite{hirohata2008identifying}], [\cite{kim2011automatic}]. CRF model employed structural and sequential information along with the lexical and semantic features and achieved significant scores on classifying structured abstracts with five classes. 

Aforementioned work proposed shallow models to classify sentences of biomedical abstracts. Along with the increase in size of structured biomedical abstract corpus, Neural Network based models have been applied to automatically learn the features based on character and word embeddings. Among CNN based models, [\cite{kim2014convolutional}] proposed a one layer CNN on top of embedding vectors with the task independent classification purposes. Despite its success in many classification tasks, the proposed model only considers N-grams and neglects long-term dependencies between words. Many studies have augmented the task-specific features with neural networks to further tune their models.  
[\cite{ma2015dependency}] proposed dependency-based convolution model that respects linguistic structures in the text. 

RNN-CRF based models capture the long-term dependencies in sequential data. They are efficiently used to embed short texts for short-text classification in [\cite{lee2016sequential}] and for sectioning of biomedical abstracts in [\cite{dernoncourt2016neural}]. These models adopt a hybrid of character and word embeddings and learned a LSTM encoder over them to embed a sentence. They also utilized CRF layer on the top layer to jointly classify subsequent sentences. More recently, [\cite{jin2018hierarchical}] employed two layers of LSTM at sentence and document level which mimics the hierarchical structure of a document. In the same fashion, they used CRF on the top layer to optimize predicted label sequences. Their approach achieved highly successful results and currently is the state-of-the-art for sequential sentence classification with biomedical abstracts.

\section{Methods}
\label{sec:methods}

\subsection{Architecture of SSN-4 Model}
\begin{figure}[h]
    \centering
    \includegraphics[width=0.8\textwidth]{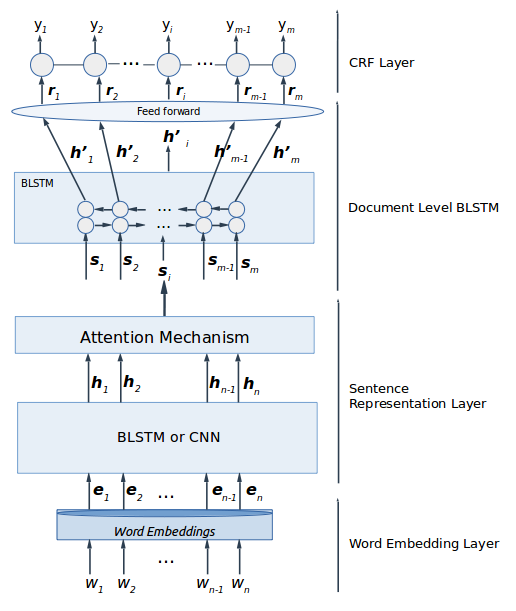} % second figure itself
    \caption{SSN-4 model for sequential sentence classification.
    $w$: word, $\mathbf{e}$: word embedding vector, $\mathbf{h}$: sentence level BLSTM/CNN output, $\mathbf{s}$: sentence encoding vector, $\mathbf{h^{'}}$: document level BLSTM output $\mathbf{r}_j$: probability vector for labels, $y_j$: sentence label.}
   \label{Figure_1}
\end{figure}

In this section, we describe the SSN-4 model proposed in [\cite{jin2018hierarchical}]. There are four layers stacked in the SSN-4 model. These layers are named from bottom (input) to top (output): word embedding layer, sentence representation layer that includes a BLSTM or CNN layer and an attention mechanism, document level BLSTM layer and CRF layer. In the following each of these components is explained in great detail. Figure \ref{Figure_1} shows the architecture of the network.

\subsubsection{Word Embedding Layer}
Given a word $w$, word embedding layer maps each word $w$ to a vector $\mathbf{e}$ in a d-dimensional word embedding space where $\mathbf{e} \in \mathbb{R}^d$. Each word has a corresponding embedding in the word embedding space that captures its lexical semantic representation. Word embeddings are pretrained using the Word2Vec model [\cite{mikolov2013efficient}]. 

\subsubsection{Sentence Representation Layer}
This layer takes as input the sequence of embedding vectors $\mathbf{e}_{1:n}$ representing a sequence of words $w_{1:n}$ in a given sentence and feeds them to an encoder network.
As suggested in [\cite{jin2018hierarchical}], CNN or BLSTM networks can be configured as a sentence encoder. We used BLSTM as it is part of the model that gives the state-of-the-art results. BLSTM structures two sets of neurons in two time directions to get information from past and future hidden states. These hidden states are concatenated at each hidden unit $i$ corresponding to word $w_i$. Because at each word position, we are receiving information from both past and future words, BLSTM can learn the complex lexical and semantic associations between words. BLSTM network outputs a sequence of hidden states $\mathbf{h}_{1:n}$ for a sentence with $n$ words with each hidden state representing the word at position $i$.

SSN-4 utilizes an attention mechanism at the word level called self-attendance [\cite{lin2017structured}]. The purpose of this attention model is to learn multiple sentence embeddings, each obtained by attending over a distinctive aspect of a sentence. We have a sequence of BLSTM hidden states $\mathbf{h}_{1:n}$, each corresponding a word in a sentence with $n$ words. We let matrix $\mathbf{H}^{u\textrm{x}n}$ be the concatenation of those hidden states, where $u$ is the dimension of the BLSTM hidden states at each position. We have two matrices defining attention model's parameters. $\mathbf{W}_{m1}$ is $a$ by $u$ matrix which transforms hidden unit outputs to another vector space with dimension $a$ where $a$ is a hyper-parameter. Then another matrix $\mathbf{W}_{m2}$, whose dimension is $r$ by $a$ is multiplied with the activation vectors formed in the previous step. Softmax is applied at each row of the resulting matrix to normalize these activations. Here the hyper-parameter $r$ is ideally the number of components, aspects planned to learn over a sentence.   

\begin{equation}
    \mathbf{A} = softmax((\mathbf{W}_{m2}tanh((\mathbf{W}_{m1}\mathbf{H})_{a\textrm{x}n}))_{r\textrm{x}n})
\end{equation}

\begin{equation}
    \mathbf{S} = \mathbf{A}\mathbf{H}
\end{equation}

Effectively, by multiplying each row of the matrix $\mathbf{A}$ with $\mathbf{H}$, we obtain a weighted linear combination of the hidden units. Sentence encoding vector $\mathbf{s}_i$ for the sentence $i$ is constituted by flattening rows of $\mathbf{S}$ into a vector.

\subsubsection{Document Level BLSTM Layer}

This layer takes as input a sequence of sentence encoders $\mathbf{s}_{1:m}$ corresponding to the sentences in a given abstract and outputs a sequence of probability vectors $\mathbf{r}_{1:m}$. 
Structure of the BLSTM is equivalent to the one utilized in the previous layer. Each hidden unit output is given as input to a feed-forward neural network with one hidden layer which subsequently gives the probability vector $\mathbf{r}$. Each element $k$ of this vector corresponds to the probability that given sentence belongs to the $k^{th}$ category.   

\subsubsection{CRF Layer}
Using a linear statistical model on the top layer aims to capture implicit patterns in the logical structure of an abstract. This structure is helpful in case observing sequential features helps discriminating sentences.
This layer takes as input a sequence of probability vectors $\mathbf{r}_{1:m}$ from the previous layer for a sequence of $m$ sentences in a given abstract and outputs a sequence of labels $y_{1:m}$ where $y_i$ is what model thinks is the correct category for the sentence. CRF is used to model the dependencies between consecutive sentence categories. To model dependencies, a transition matrix $\mathbf{T}$ is learned using the categories of consecutive sentences in the abstract data set. $\mathbf{T}_{(i,j)}$ represents the probability that a sentence with category j follows the sentence with category i in a given abstract. In order to jointly predict sequence categories, a score function is assigned to each sequence of categories. Scores can be turned into probabilities by applying softmax function over scores of all category sequences.

\begin{equation}
    s(y_{1:m}) = \sum_{i=1}^{m}\mathbf{r}_i + \sum_{i=2}^{m}\mathbf{T}(y_{i-1},y_i)
\end{equation}

Throughout the training phase the objective is to maximize the log probability of the ground truth sequence. In the test phase, Dynamic Programming based Viterbi algorithm [\cite{forney1973viterbi}] is used to choose the best sequence.

\vspace{0.5cm}

\subsection{Data sets}
We use the PubMed 200k RCT dataset  [\cite{dernoncourt2017pubmed}] for our evaluation purposes. This dataset consists of approximately 200k biomedical abstracts explicitly selected among Randomized Controlled Trials (RCT) studies and collected from Medline/PubMed database. These abstracts are structured, meaning that each sentence of an abstract has tagged with a semantic role using the following : Background, Objective, Method, Result and Conclusion.

\vspace{0.5cm}

\begin{table}[h]
    %\hspace{0.5cm}
    %\begin{minipage}{0.5\textwidth}
        \centering
        %\resizebox{\columnwidth}{!}{
        \begin{tabular}{|l|c|c|c|c|} % <-- Changed to S here.
         \hline
        Dataset & $\lvert V \rvert$ & Train & Validation & Test\\
        \hline
      PubMed 200k & 331k & 190k (2.2M) & 2.5k (29k) & 2.5k (30k)\\
      \hline
    \end{tabular}
    \vspace{0.1in}
    \caption{Overview of PubMed 200k RCT dataset. V represents the vocabulary set. For the training, validation and test sets, the number of abstracts are followed by the total number of sentences.}
    \label{tab:data}
    \end{table}

\subsection{Experiments}
\subsubsection{Contribution of Attention Mechanism and CRF Layer}
It's unclear whether the individual components of the SSN-4 model have significant contribution because they create complex neural paths when they are layered together. The model, therefore could overfit. We need to study components individually to observe if they are useful at the expense of increasing complexity. 

\hspace*{0.8cm} a) {\it{Contribution of Attention Mechanism: }} As pointed out, while defining the model, attention used in the SSN-4 model brings extra complexity in the sentence embedding layer by training two additional matrices and production of $r$ different sentence encoders for each sentence. To observe its performance contribution, we first discard the attention weight matrix. Instead we extract the output of the final hidden unit in the forward and backward LSTM pass and concatenate them to use it as a sentence embedding. In this way, we bypass the attention paths and let all the information flow through BLSTM. We will call this model SSN-4-Attention.  

\hspace*{0.8cm} b) {\it{Contribution of CRF Layer: }} CRF learns transition probabilities between subsequent categories. However, as a side effect of being at the top layer, its gradient updates propagates to all the previous layers during back-propagation. This brings another level of unpredictability in the model. Therefore, we need to investigate its effect on the performance of the model. To achieve that we eliminate CRF, hence opt out to learn the transition probabilities jointly with the rest of the model. We instead impose rules on the sequence and the position of the predicted layers. For instance, we say Conclusion appears later in the sentence and is not followed by any other category. During the decoding most likely sequence is selected following these rules. We will call this model SSN-4-CRF. 

\subsubsection{How Well SSN-4 Model Generalizes to Other Dataset}
In this experiment, we explore whether the SSN-4 model generalizes to a new data set. In [\cite{jin2018hierarchical}] the model is trained and tested with author's annotated set which consists of Randomized Controlled Trials. RCT is a special type of study that follows common category sequences and language patterns. We want to apply the model on any abstract whether or not it discusses Randomized Trials.

\hspace*{0.8cm} a) {\it{Collection of Dataset: }} We collect new abstracts from Medline database that have total of 1.6 M abstracts with 19 M sentences. These are structured abstracts and have already been divided into sections. The authors' division into sections have been mapped into our 5 categories by NLM. We reference this set as Medline Diverse Set (MDS). During the collection process we don't filter abstracts by the year of the study or the type of the study. We make sure that abstracts have the same five categories and and they are well-formatted. We select 200k abstracts randomly from this set and use it as MDS test set.

\hspace*{0.8cm} b) {\it{Evaluation: }} The SSN-4 Model trained on MDS will be called SSN-4-MDS. Its performance on RCT test set \& MDS test set will be compared to SSN-4 models' performance on the same test sets. 

\subsubsection{Fine-tuning Word Embeddings}

SSN-4 Model uses generic lexical representation pre-trained on Word2Vec [\cite{mikolov2013efficient}].
Following the idea presented in [\cite{wang2015predicting}], by tailoring the word embeddings to the task, task-specific representation can be learned. 
By applying the methods below we compare the performance of the end-to-end model under three scenarios.
\begin{itemize}
\item[-] We use Word2Vec 200-dimensional embeddings.
\item[-] We further tune Word2Vec embeddings jointly. We call the resulting model SSN-4-WEFT.
\item[-] We randomly initialize word embeddings  as described in [\cite{glorot2010understanding}] and learn them during training. We call the resulting model SSN-4-WERI.
\end{itemize}

\subsection{Building a Network for Prediction Correction}

In this experiment, we take a simple method to correct predictions of the SSN-4-CRF model on the most confused category pair. As previously discussed, SSN-4-CRF outputs a probability vector $\mathbf{v}_i$ per sentence where each element of this vector represents a prediction score for one of the five categories. We accept a score vector whenever SSN-4-CRF predicts either one of the confused pair and we subsequently input that to MLP-2 model. We prepare a validation set from the MDS dataset to extract the predictions of the SSN-4-CRF model.  

We use $\mathbf{v}_i$ as the input features and train the MLP-2 model. We have three output classes in the MLP-2 model. Two of them are for the confused category pair and the third one is "Other". The Other class is used to map to any one of the other three categories in our original 5-class set, whenever the correct category is among those three but SSN-4-CRF mistakenly predicts it either one of the confused pair.

\begin{figure}[h]
    \centering
    \includegraphics[width=0.9\textwidth]{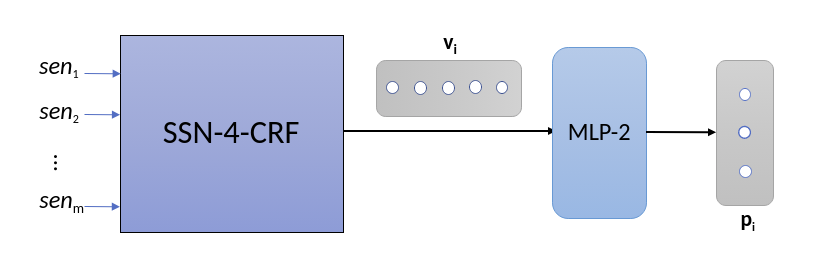} % second figure itself
    \caption{Model pipeline. $\mathbf{v}_i$ is the score vector predicted by SSN-4-CRF for each of the five categories. $\mathbf{p}_i$ is the score vector predicted by MLP-2 for each of the three categories.}
   \label{Figure_2}
\end{figure}

\section{Results \& Discussion}
%\begin{comment}
\subsection{Performance of SSN-4 Model}

The precision, recall and F1-score of the SSN-4 model on the RCT test set are \%93.91, \%93.92, \%93.91 respectively. To further analyse the performance we construct the confusion matrix which is shown in Table \ref{tab:confusion}. The main confusion appears between the Background and Objective categories that critically drops the scores.
\vspace{0.8cm}
\begin{table}[h]
    %\hspace*{-0.5cm}
    %\begin{minipage}{0.5\textwidth}
        \centering
        \begin{tabular}{|l|c|c|c|c|c|} % <-- Changed to S here.
         \hline
      \backslashbox{Actual}{Prediction}  & Backg. & Concl. & Mtds. & Obj. & Res.\\
        \hline
      Background & 1999 & 0 & 50 & 419 & 0\\
      \hline
      Conclusion & 6 & 4269 & 11 & 0 & 140\\
      \hline
      Methods & 27 & 10 & 9551 & 7 & 143\\
      \hline
      Objectives & 544 & 0 & 53 & 1778 & 1 \\
      \hline
      Results & 7 & 76 & 286 & 0 & 9892 \\
      \hline
    \end{tabular}
    \vspace{0.1in}
    \caption{Confusion matrix of the SSN-4 model trained on the RCT data set. Rows corresponding to the actual label while columns shows the predicted labels. }
    \label{tab:confusion}
   % \end{minipage}
    %\hfillx
    \end{table}

\subsection{Contribution of Attention and CRF Layers}

As discussed in the previous section, we developed two new models called SSN-4-CRF and SSN-4-Attention to analyse the contribution of the CRF and Attention layers in the SSN-4 model. Here we will show the precision, recall and F-1 scores of these models and discuss their performance. 

SSN-4-CRF: The SSN-4 model performs slightly better than the SSN-4-CRF model. As can be seen in Table \ref{tab:attention}, there is a drop of 1\% in both precision and recall, showing that CRF's transition probabilities helps in modeling the transitions and that just imposing an order constraint as in SSN-4-CRF is not sufficient.

For a closer examination we looked at the confusion matrix and observed that the original model performs better at capturing transitions between neighbour categories especially between Background and Objective or Methods and Results. On the other hand, imposing the order constraint in the model helped prevent the few cases of predicting Background after Methods or Methods after Conclusions.
  
SSN-4-Attention: Table \ref{tab:attention} shows that even without the complex attention layer, the new model performs almost as well as the original model. In the future, we could consider whether the extra parameters that needed to be learned in the attention model can cause overfitting. Examination of the confusion matrix didn't show any significant trends. 

We also developed another model that has neither attention nor CRF layers that is shown in the Table \ref{tab:attention}. Since the results are very similar to SSN-4-CRF, once more we see that attention is really not adding much to improve the results. Our overall conclusion is that CRF seems to be important for the performance of the SSN-4 model.  

%\vspace{0.3cm}
\begin{table}[h!]
\sisetup{round-mode          = places, % Rounds numbers
  round-precision     = 2, % to 2 places
}
\centering
  \begin{center}
    \label{tab:table1}
    %\resizebox{\columnwidth}{!}{
    \begin{tabular}{lSSS} % <-- Changed to S here.
      \toprule
       Model & \multicolumn{1}{c}{Precision} & \multicolumn{1}{c}{Recall} & \multicolumn{1}{c}{F1-Score} \\
      \midrule
      Attention + , CRF +  & 93.91 & 93.92 & 93.91 \\
      Attention + , CRF \hspace{0.05cm}-  & 93.04 & 93.04 & 92.96 \\
      Attention \hspace{0.05cm}- \hspace{0.05cm}, CRF +  & 93.69 & 93.61 &  93.55\\
      Attention  \hspace{0.05cm}- \hspace{0.05cm}, CRF \hspace{0.05cm}-  & 92.96 & 92.93 & 92.88 \\
      \bottomrule
    \end{tabular}
    %}
    \vspace{0.1in}
    \caption{Row 1 corresponds to SSN-4 , row 2 corresponds to SSN-4-CRF, row 3 corresponds to SSN-4-Attention, and Row 4 corresponds to a model without CRF or Attention.}
    \label{tab:attention}
  \end{center}
\end{table}

\vspace{-0.2cm}
\subsection{Model Generalization}
The first two rows of the Table \ref{tab:general} shows that SSN-4 model does not appear to generalize well beyond Randomized Controlled Trials subset, as can be seen from the significant drop off seen in the second row of the table. The next two rows show the results of the model trained on the much larger MDS set (SSN-4-MDS). The performance of this model is slightly worse than SSN-4 when tested on RCT test set. However, the new model seems to perform almost as well in both RCT and the more general domain test set. These results seem to be more evidence of the overfitting of the SSN-4 model to the Randomized Controlled Trials domain. 

\vspace{0.3cm}
\begin{table}[h!]
\sisetup{round-mode          = places, % Rounds numbers
  round-precision     = 2, % to 2 places
}
\centering
  \begin{center}
    \label{tab:table4}
    %\resizebox{\columnwidth}{!}{
    \begin{tabular}{lSSS} % <-- Changed to S here.
      \toprule
       Model & \multicolumn{1}{c}{Precision} & \multicolumn{1}{c}{Recall} & \multicolumn{1}{c}{F1-Score} \\
      \midrule
      SSN-4, tested on RCT  & 93.91 & 93.92 & 93.91 \\
      SSN-4, tested on MDS  & 88.35 & 87.91 & 87.79 \\
      SSN-4-MDS, tested on RCT  & 92.32 & 92.34 &  92.29\\
      SSN-4-MDS, tested on MDS  & 91.78 & 91.75 & 91.73 \\
      \bottomrule
    \end{tabular}
    %}
    \vspace{0.1in}
    \caption{Rows 1 and 2 provide the results of the SSN-4 model. Rows 3 and 4 provide the results of the SSN-4-MDS.}
    \label{tab:general}
  \end{center}
\end{table}

\vspace{-0.5cm}
\subsection{Fine-tuning Word Embeddings}

SSN-4 model uses the general word embedding model trained by Word2Vec. There is a possibility that these word embeddings don't differentiate between different contiguous sections. Like with the previous work we conducted a study to see if the word embeddings could be fine-tuned for the task. However,
there is no significant change overall between the scores of the pre-trained and fine-tuned models  as seen in Table \ref{tab:weft}. To see if the fine-tuned model make any difference for a specific section we examined the results for each section.
We observed that there is hardly any noticeable differences.

 Rather than fine tune an already good quality word embeddings, we also considered tuning the word embeddings starting with the random initialization. Not surprisingly, the results are worse than starting with the pre-trained word embeddings. However, since the results are still within 2\% percent of the other two results, it might be interesting to see how the results would change if we attempted to fine tune the word embeddings using a much larger training set such as MDS.

\vspace{0.3cm}
\begin{table}[h!]
\sisetup{round-mode          = places, % Rounds numbers
  round-precision     = 2, % to 2 places
}
\centering
  \begin{center}
    \label{tab:table5}
    %\resizebox{\columnwidth}{!}{
    \begin{tabular}{lSSS} % <-- Changed to S here.
      \toprule
       Model & \multicolumn{1}{c}{Precision} & \multicolumn{1}{c}{Recall} & \multicolumn{1}{c}{F1-Score} \\
      \midrule
      SSN-4 & 93.91 & 93.92 & 93.91 \\
      SSN-4-WEFT & 93.61 & 93.63 &  93.61\\
      SSN-4-WERI & 92.18 & 92.21 & 92.11 \\     \bottomrule
    \end{tabular}
    %}
    \vspace{0.1in}
    \caption{Results with different choices of word embeddings on the sequential sentence classification task.}
    \label{tab:weft}
  \end{center}
\end{table}
%\end{comment}
\vspace{-0.4cm}
\subsection{Building a Network for Prediction Correction}
Although SSN-4 model gives significant results, we want to explore where there is a problem in order to see if further improvement might be possible. As revealed in the analysis of the model, the confusion happens mostly between the Background and Objective categories. Therefore, we used these two categories in the previously discussed network as the most confused pair. Table \ref{tab:mlpprec}, \ref{tab:ssnprec} show that SSN-CRF + MLP-2 has a marginal improvement in F-1 score. This result is somewhat encouraging. As noted, our method represents probably the simplest approach to improve among the confusion classes prediction. In the future we can attempt to develop a more complex method, that considers more aspects of the input (words, sentences \& sequence of sentences) instead of just the output of SSN-4-CRF.
\vspace{0.5cm}
\begin{table}[h]
    \hspace*{-0.5cm}
    \begin{minipage}{0.5\textwidth}
        \centering
        \resizebox{\columnwidth}{!}{
        \begin{tabular}{|l|c|c|c|} % <-- Changed to S here.
         \hline
       \backslashbox{Category}{Measure} & Precision & Recall & F1-Score\\
        \hline
      Background & 75.96 & 83.33 & 79.48\\
      \hline
      Objective & 80.25 & 73.12 & 76.52\\
      \hline
    \end{tabular}}
    \vspace{0.1in}
    \caption{Precision, recall and F1-Score by SSN-4-CRF + MLP-2 for the Background and Objective categories.}
    \label{tab:mlpprec}
    \end{minipage}
    %\hfillx
    \hspace*{0.8cm}
    \begin{minipage}{0.5\textwidth}
    \centering
    \resizebox{\columnwidth}{!}{
    \begin{tabular}{|l|c|c|c|} % <-- Changed to S here.
         \hline
        \backslashbox{Category}{Measure} & Precision & Recall & F1-Score\\
        \hline
      Background & 74.15 & 85.01 & 79.21\\
      \hline
      Objective & 81.33 & 70.02 & 75.24\\
      \hline
    \end{tabular}}
    \vspace{0.1in}
    \caption{Precision, recall and F1-Score by SSN-4-CRF for the Background and Objective categories.}
    \label{tab:ssnprec}
   % \label{tab:tab6}
     \end{minipage}
\end{table}
\section{Conclusion}

We first examined the contribution of the Attention and CRF Layers of the SSN-4 model. Our approach uses BIOBERT based model, with a moving window to contextualize the abstracts. Experiment results show that just imposing the order constraint in the sentence categories is not sufficient to replace the CRF layer. On the other hand, without the attention layer, SSN-4 model performs nearly as well.

We collected a new data set we called MDS that is considerably bigger than PubMed RCT dataset and includes abstracts that are not limited to RCT. We show that SSN-4 model does not appear to generalize beyond RCT dataset. However the model trained with the MDS performs fairly well in both MDS and RCT data sets.

We explored whether lexical representations can be tailored to the given task. We first fine-tuned the words embeddings starting from the embeddings obtained from Word2Vec. While we didn't see any significant changes, fine-tuning word embeddings starting from random initialization results in drop in the scores.  

In the future, we would like to do more work using the MDS data set, considering that the model trained with this set generalizes better in different domains therefore, prove that it is more valuable in general biomedical domain. We plan to fine-tune word embeddings on this model to see if having a much larger training set makes a difference. We plan to do extended experiments with a new model to improve on the confused pair of categories.

%\bibliographystyle{unsrtnat}
%\bibliography{references}  %%% Uncomment this line and comment out the ``thebibliography'' section below to use the external .bib file (using bibtex) .





% \end{thebibliography}

\printbibliography

@article{jin2018hierarchical,
  title={Hierarchical Neural Networks for Sequential Sentence Classification in Medical Scientific Abstracts},
  author={Jin, Di and Szolovits, Peter},
  journal={arXiv preprint arXiv:1808.06161},
  year={2018}
}

@article{lee2016sequential,
  title={Sequential short-text classification with recurrent and convolutional neural networks},
  author={Lee, Ji Young and Dernoncourt, Franck},
  journal={arXiv preprint arXiv:1603.03827},
  year={2016}
}

@article{dernoncourt2016neural,
  title={Neural networks for joint sentence classification in medical paper abstracts},
  author={Dernoncourt, Franck and Lee, Ji Young and Szolovits, Peter},
  journal={arXiv preprint arXiv:1612.05251},
  year={2016}
}

@article{dernoncourt2017pubmed,
  title={Pubmed 200k rct: a dataset for sequential sentence classification in medical abstracts},
  author={Dernoncourt, Franck and Lee, Ji Young},
  journal={arXiv preprint arXiv:1710.06071},
  year={2017}
}

@article{mikolov2013efficient,
  title={Efficient estimation of word representations in vector space},
  author={Mikolov, Tomas and Chen, Kai and Corrado, Greg and Dean, Jeffrey},
  journal={arXiv preprint arXiv:1301.3781},
  year={2013}
}

@inproceedings{kim2011automatic,
  title={Automatic classification of sentences to support evidence based medicine},
  author={Kim, Su Nam and Martinez, David and Cavedon, Lawrence and Yencken, Lars},
  booktitle={BMC bioinformatics},
  volume={12},
  number={2},
  pages={S5},
  year={2011},
  organization={BioMed Central}
}

@article{ruch2007using,
  title={Using argumentation to extract key sentences from biomedical abstracts},
  author={Ruch, Patrick and Boyer, Celia and Chichester, Christine and Tbahriti, Imad and Geissb{\"u}hler, Antoine and Fabry, Paul and Gobeill, Julien and Pillet, Violaine and Rebholz-Schuhmann, Dietrich and Lovis, Christian and others},
  journal={International journal of medical informatics},
  volume={76},
  number={2-3},
  pages={195--200},
  year={2007},
  publisher={Elsevier}
}

@article{liu2013abstract,
  title={Abstract sentence classification for scientific papers based on transductive SVM},
  author={Liu, Yuanchao and Wu, Feng and Liu, Ming and Liu, Bingquan},
  journal={Computer and Information Science},
  volume={6},
  number={4},
  pages={125},
  year={2013},
  publisher={Canadian Center of Science and Education}
}

@article{forney1973viterbi,
  title={The viterbi algorithm},
  author={Forney, G David},
  journal={Proceedings of the IEEE},
  volume={61},
  number={3},
  pages={268--278},
  year={1973},
  publisher={IEEE}
}

@article{kim2014convolutional,
  title={Convolutional neural networks for sentence classification},
  author={Kim, Yoon},
  journal={arXiv preprint arXiv:1408.5882},
  year={2014}
}

@article{ma2015dependency,
  title={Dependency-based convolutional neural networks for sentence embedding},
  author={Ma, Mingbo and Huang, Liang and Xiang, Bing and Zhou, Bowen},
  journal={arXiv preprint arXiv:1507.01839},
  year={2015}
}

@inproceedings{lin2006generative,
  title={Generative content models for structural analysis of medical abstracts},
  author={Lin, Jimmy and Karakos, Damianos and Demner-Fushman, Dina and Khudanpur, Sanjeev},
  booktitle={Proceedings of the hlt-naacl bionlp workshop on linking natural language and biology},
  pages={65--72},
  year={2006},
  organization={Association for Computational Linguistics}
}

@article{lin2017structured,
  title={A structured self-attentive sentence embedding},
  author={Lin, Zhouhan and Feng, Minwei and Santos, Cicero Nogueira dos and Yu, Mo and Xiang, Bing and Zhou, Bowen and Bengio, Yoshua},
  journal={arXiv preprint arXiv:1703.03130},
  year={2017}
}

@inproceedings{wang2015predicting,
  title={Predicting polarities of tweets by composing word embeddings with long short-term memory},
  author={Wang, Xin and Liu, Yuanchao and Chengjie, SUN and Wang, Baoxun and Wang, Xiaolong},
  booktitle={Proceedings of the 53rd Annual Meeting of the Association for Computational Linguistics and the 7th International Joint Conference on Natural Language Processing (Volume 1: Long Papers)},
  volume={1},
  pages={1343--1353},
  year={2015}
}

@inproceedings{glorot2010understanding,
  title={Understanding the difficulty of training deep feedforward neural networks},
  author={Glorot, Xavier and Bengio, Yoshua},
  booktitle={Proceedings of the thirteenth international conference on artificial intelligence and statistics},
  pages={249--256},
  year={2010}
}

@inproceedings{hirohata2008identifying,
  title={Identifying sections in scientific abstracts using conditional random fields},
  author={Hirohata, Kenji and Okazaki, Naoaki and Ananiadou, Sophia and Ishizuka, Mitsuru},
  booktitle={Proceedings of the Third International Joint Conference on Natural Language Processing: Volume-I},
  year={2008}
}

\end{document}